# Explicit Learning Curves for Transduction and Application to Clustering and Compression Algorithms

**Philip Derbeko**                                           PHILIP@CS.TECHNION.AC.IL

**Ran El-Yaniv**                                              RANI@CS.TECHNION.AC.IL
*Department of Computer Science*
*Technion - Israel Institute of Technology*
*Haifa 32000, Israel*

**Ron Meir**                                                 RMEIR@EE.TECHNION.AC.IL
*Department of Electrical Engineering*
*Technion - Israel Institute of Technology*
*Haifa 32000, Israel*

## Abstract

Inductive learning is based on inferring a general rule from a finite data set and using it to label new data. In *transduction* one attempts to solve the problem of using a labeled training set to label a set of unlabeled points, which are given to the learner prior to learning. Although transduction seems at the outset to be an easier task than induction, there have not been many *provably useful* algorithms for transduction. Moreover, the precise relation between induction and transduction has not yet been determined. The main theoretical developments related to transduction were presented by Vapnik more than twenty years ago. One of Vapnik's basic results is a rather tight error bound for transductive classification based on an exact computation of the hypergeometric tail. While tight, this bound is given *implicitly* via a computational routine. Our first contribution is a somewhat looser but *explicit* characterization of a slightly extended PAC-Bayesian version of Vapnik's transductive bound. This characterization is obtained using concentration inequalities for the tail of sums of random variables obtained by sampling without replacement. We then derive error bounds for compression schemes such as (transductive) support vector machines and for transduction algorithms based on clustering. The main observation used for deriving these new error bounds and algorithms is that the unlabeled test points, which in the transductive setting are known in advance, can be used in order to construct useful *data dependent* prior distributions over the hypothesis space.

## 1. Introduction

The bulk of work in Statistical Learning Theory has dealt with the *inductive* approach to learning. Here one is given a finite set of labeled training examples, from which a rule is inferred. This rule is then used to label new examples. As pointed out by Vapnik (1998) in many realistic situations one actually faces an easier problem where one is given a training set of labeled examples, together with an unlabeled set of points which needs to be labeled. In this *transductive* setting, one is not interested in inferring a general rule, but rather only in labeling this unlabeled set as accurately as possible. One solution is of course to infer a rule as in the inductive setting, and then use it to label the required points. However, as argued by Vapnik (1982, 1998), it makes little sense to solve what appears to be an easier problem by 'reducing' it to a more difficult one. While there are currently no formal results stating that transduction is indeed easier than induction[1] it is plausible that the relevant information carried by the test points can be incorporated into an algorithm, potentially leading to superior performance. Since in many practical situations we are interested in evaluating a function only at some points of interest, a major open problem in statistical learning theory is to determine precise relations between induction and transduction.

In this paper we present several general error bounds for transductive learning.[2] We also present a general technique for establishing error bounds for transductive learning algorithms based on compression and clustering. Our bounds can be viewed as extensions of McAllester's PAC-Bayesian framework (McAllester, 1999, 2003a, 2003b) to transductive learning. The main advantage of using the PAC-Bayesian approach in transduction, as opposed to induction, is that here prior beliefs on hypotheses can be formed *based on the unlabeled test data*. This flexibility allows for the choice of "compact priors" (with small support) and therefore, for tight bounds. We use the established bounds and provide tight error bounds for "compression schemes" such as (transductive) SVMs and transductive learning algorithms based on clustering. While precise relations between induction and transduction remain a major challenge, our new bounds and technique offer some new insights into transductive learning.

The problem of transduction was formulated as long ago as 1982 in Vapnik's classic book (Vapnik, 1982), where the precise setting was formulated, and some implicit error bounds were derived.[3] In recent years the problem has been receiving an increasing amount of attention, due to its applicability to many real world situations. A non-exhaustive list of recent contributions includes (Vapnik, 1982, 1998; Joachims, 1999; Bennett & Demiriz, 1998; Demiriz & Bennett, 2000; Wu, Bennett, Cristianini, & Shawe-Taylor, 1999; Lanckriet, Cristianini, Ghaoui, Bartlett, & Jordan, 2002), and (Blum & Langford, 2003). Most of this work, with the exception of Vapnik's bounds (1982, 1998) and the results of Lanckriet et al. (2002), has dealt with algorithmic issues, rather than with performance bounds. Implicit performance bound, in the spirit of Vapnik (1982, 1998) has recently been presented by Blum and Langford (2003). We mention this result again later in Section 2.2.

We present *explicit* PAC-Bayesian bounds for a transductive setting that considers sampling *without replacement* of the training set from a given 'full sample' of unlabeled points. This setting is proposed by Vapnik and it turns out that error bounds for learning algorithms, within this setting, imply the same bounds within another setting which may appear more practical (See Section 2.1 and Theorem 2 for details). Sampling without replacement of the training set leads to the training points being *dependent* (see Section 2.1 for details). Our first goal is to provide uniform bounds on the deviation between the training error and the test error. To this end, we study two types of bounds that utilize two different bounding techniques. The first approach is based on an observation made in Hoeffding's classic paper (Hoeffding, 1963). This approach was recently alluded to by Lugosi (2003). As pointed out by Hoeffding (1963), the sampling without replacement problem can be reduced to an equivalent problem involving independent samples, for which standard techniques suffice. We refer to this approach as 'reduction to independence'. A second approach involves derivations of large deviation bounds for sampling *without replacement* such as those developed by Serfling (1974), Vapnik (1998), Hush and Scovel (2003), and Dembo and Zeitouni (1998). We refer to such bounds as 'direct'. We consider these two approaches in Section 3.1 and 3.2, respectively. Using these two approaches we derive general PAC-Bayesian bounds for transduction. It turns out that the direct bounds lead to tighter and more explicit learning curves for transduction.

---

1. There may be various ways to state this in a meaningful manner. Essentially, we would like to know if for some learning problems a particular transductive algorithm can achieve better performance than any inductive algorithm, where performance may be characterized by learning rates and/or computational complexity.

2. In the paper we use the terms 'error bound' and 'risk bound' interchangeably. Strictly speaking, the term 'generalization bound' is not appropriate in the transductive setting.

3. Vapnik refers to transduction also as "estimating the values of a function at given points of interest".

We then show how to utilize PAC-Bayesian transductive bounds to derive error bounds for specific learning schemes. In particular, we show how to choose priors, based on the given unlabeled data, and derive bounds for "compression schemes" and learning algorithms based on clustering. Compression schemes are algorithms that select the same hypothesis using only a subset of the training data. The main example of a compression scheme is a (transductive) Support Vector Machine (SVM). The compression level achieved by such schemes typically depends on the (geometric) complexity of the learning problem at hand, which sometimes does not allow for significant compression. A stronger type of compression can be often achieved using clustering. A natural approach in the context of transduction (and semi-supervised learning) is to apply a clustering algorithm over the set of all available (unlabeled) points and then to use the labeled points to determine the classifications of points in the resulting clusters. We formulate this scheme in the context of transduction and derive for it rather tight error bounds by utilizing an appropriate prior in our transductive PAC-Bayesian bounds. For practical applications a similar but tighter result is obtained by using the implicit bounds of Vapnik (1998) (or the bound of Blum & Langford, 2003).

## 2. Problem Setup and Vapnik's Basic Results

### 2.1 Problem Setup

The problem of transduction can be informally described as follows. A learner is given a set of labeled examples $\{(x_1, y_1), \ldots, (x_m, y_m)\}$, $x_i \in \mathcal{X}$, $y_i \in \mathcal{Y}$, and a set of unlabeled points $\{x_{m+1}, \ldots, x_{m+u}\}$. Based on this data, the objective is to label the unlabeled points. In order to formalize the scenario, we consider the setting proposed by Vapnik (1998, Chap. 8), which for simplicity is described in the context of binary classification.[4] Let $\mathcal{H}$ be a set of binary hypotheses consisting of functions from the input space $\mathcal{X}$ to $\mathcal{Y} = \{\pm 1\}$. Let $\mu(x, y)$ be any distribution over $\mathcal{X} \times \mathcal{Y}$. For each $h \in \mathcal{H}$ and a set $Z = x_1, \ldots, x_{|Z|}$ of samples define

$$R_h(Z) \triangleq \frac{1}{|Z|} \sum_{i=1}^{|Z|} \mathbf{E}_{\mu(y|x_i)} \{\ell(h(x_i), y)\} = \frac{1}{|Z|} \sum_{i=1}^{|Z|} \int_{y \in \mathcal{Y}} \ell(h(x_i), y) d\mu(y|x_i), \quad (1)$$

where, unless otherwise specified, $\ell(\cdot, \cdot)$ is the 0/1 loss function. Vapnik (1998) considers the following two transductive "protocols".

---

4. Note that the bound of Theorem 22 does hold for general bounded loss functions.

***Setting 1***:

(i) A full sample $X_{m+u} = \{x_1, \ldots, x_{m+u}\}$ consisting of arbitrary $m + u$ points is given.[5]

(ii) We then choose uniformly at random the training sample $X_m \subseteq X_{m+u}$ and receive its labels $Y_m$ where the label $y$ of $x$ is chosen according to $\mu(y|x)$; the resulting training set is $S_m = (X_m, Y_m)$ and the remaining set $X_u$ is the *unlabeled (test) sample*, $X_u = X_{m+u} \setminus X_m$.

(iii) Using both $S_m$ and $X_u$ we select a classifier $h \in \mathcal{H}$. The quality of $h$ is measured by $R_h(X_u)$.

***Setting 2***:

(i) We are given a training set $S_m = (X_m, Y_m)$ selected i.i.d according to $\mu(x, y)$.

(ii) An independent test set $S_u = (X_u, Y_u)$ of $u$ samples is then selected in the same manner.

(iii) We are required to choose our best $h \in \mathcal{H}$ based on $S_m$ and $X_u$ so as to minimize

$$R_{m,u}(h) \triangleq \int \frac{1}{u} \sum_{i=m+1}^{m+u} \ell\left(h(x_i), y_i\right) d\mu(x_1, y_1) \cdots d\mu(x_{m+u}, y_{m+u}). \tag{2}$$

**Remark 1** *Notice that the choice of the sub-sample $X_m$ in Setting 1 is equivalent to sampling $m$ points from $X_{m+u}$ uniformly at random without replacement. This leads to the samples being dependent. Also, Setting 1 concerns an "individual sample" $X_{m+u}$ and there are no assumptions regarding its underlying source. The only element of chance in this setting is the random choice of the training set from the full sample.*

Setting 2 may appear more applicable in some practical situations than Setting 1. However, derivation of theoretical results can be easier within Setting 1. The following useful theorem (Theorem 8.1 in Vapnik, 1998) relates the two transduction settings. For completeness we present Vapnik's proof in the appendix.

**Theorem 2 (Vapnik)** *If for some learning algorithm choosing an hypothesis $h$ it is proved within Setting 1 that with probability at least $1 - \delta$, the deviation between the risks $R_h(X_m)$ and $R_h(X_u)$ does not depend on the composition of the full sample and does not exceed $\varepsilon$, then with the same probability, in Setting 2 the deviation between $R_h(X_m)$ and the risk given by formula (2) does not exceed $\varepsilon$.*

**Remark 3** *The learning algorithm in Theorem 2 is implicitly assumed to be deterministic. The theorem can be extended straightforwardly to the case where the algorithm is randomized and chooses an hypothesis $h \in \mathcal{H}$ randomly, based on $S_m \cup X_u$. A particular type of randomization is the one used by Gibbs algorithms, as discussed in Section 4.1.*

---

5. The original Setting 1, as proposed by Vapnik, discusses a full sample whose points are chosen independently at random according to some source distribution $\mu(x)$.

**Remark 4** *The basic quantity of interest in Setting 2 is $R_{m,u}(h)$ defined in (2). Assuming $h$ is selected based on the sample $S_m \cup X_u$, one is often interested in its expectation over a random selection of this sample. In inductive learning, one considers the sample $S_m$ and a single new point $(x,y)$, and the average is taken with respect to $S_m \cup \{X\}$. It should be noted that the random variable $R_{m,u}(h)$ is 'more concentrated' around its mean than the random variable $\ell(h(x),y)$ corresponding to a single new point $(x,y)$. Therefore, one may expect transduction to lead to tighter bounds in Setting 2 as well.*

In view of Theorem 2 we restrict ourselves in the sequel to Setting 1. Also, for simplicity we focus on the case where there exists a *deterministic* target function $\phi : \mathcal{X} \to \mathcal{Y}$, so that $y = \phi(x)$ is a fixed target label for $x$; that is, $\mu(\phi(x)|x) = 1$ (there is no requirement that $\phi \in \mathcal{H}$). Note that it is possible to extend our results to the the general case of stochastic targets $y \sim \mu(y|x)$.

We make use of the following quantities, which are all instances of (1). The quantity $R_h(X_{m+u})$ is called the *full sample risk* of the hypothesis $h$, $R_h(X_u)$ is referred to as the *transduction risk* (of $h$), and $R_h(X_m)$ is the *training error* (of $h$). Note that in the case where our target function $\phi$ is deterministic,

$$R_h(X_m) = \frac{1}{m}\sum_{i=1}^{m}\mathbf{E}_{\mu(y|x_i)}\left\{\ell(h(x_i),y)\right\} = \frac{1}{m}\sum_{i=1}^{m}\ell(h(x_i),\phi(x_i)).$$

Thus, $R_h(X_m)$ is the standard training error denoted interchangeably by $\hat{R}_h(S_m)$. It is important to observe that while $R_h(X_{m+u})$ is *not* a random variable, both $R_h(X_m)$ and $R_h(X_u)$ are random variables, due to the random selection of the samples $X_m$ from $X_{m+u}$.

While our objective in transduction is to achieve small error over the unlabeled sample (i.e. to minimize $R_h(X_u)$), we find that it is sometimes easier to derive error bounds for the full sample risk. The following simple lemma translates an error bound on $R_h(X_{m+u})$, the full sample risk, to an error bound on the transduction risk $R_h(X_u)$.

**Lemma 5** *For any $h \in \mathcal{H}$ and any $C$*

$$R_h(X_{m+u}) \le \hat{R}_h(S_m) + C \quad \Leftrightarrow \quad R_h(X_u) \le \hat{R}_h(S_m) + \frac{m+u}{u}\cdot C. \tag{3}$$

**Proof** For any $h$

$$R_h(X_{m+u}) = \frac{mR_h(X_m) + uR_h(X_u)}{m+u}. \tag{4}$$

Substituting $\hat{R}_h(S_m)$ for $R_h(X_m)$ in (4) and then substituting the result for the left-hand side of (3) we get

$$R_h(X_{m+u}) = \frac{m\hat{R}_h(S_m) + uR_h(X_u)}{m+u} \le \hat{R}_h(S_m) + C.$$

The equivalence (3) is now obtained by isolating $R_h(X_u)$ on the left-hand side. ∎

**Remark 6** *In applications of Lemma 5, the term $C = C(m)$ is typically a function $m$ (and of some of the other problem parameters such as $\delta$). In meaningful bounds $C(m) \to 0$ with $m \to \infty$. Observe that in order for the bound on $R_h(X_u)$ to converge it must be that $u = \omega(mC(m))$.*

Consider a hypothesis class $\mathcal{H}$. In the context of transduction, we are only interested in labeling the test set $X_u$, which is given in advance. Thus, we may in principle regard hypotheses in $\mathcal{H}$ which label $X_{m+u}$ identically as belonging to the same equivalence class. Since for fixed values of $m$ and $u$ the number of equivalence classes is finite (in the case of binary hypotheses, it is at most $2^{m+u}$) we may, without loss of generality, restrict ourselves to a finite hypothesis class (see, for example Vapnik, 1998, Sec. 8.5). Note that this freedom is *not* available in the inductive setting, where the test set is not known in advance.

**Remark 7** *The above procedure is clearly not possible in the case of real-valued loss functions. However, in this case one can still use the availability of the full data set $X_{m+u}$ in order to construct an empirical $\epsilon$-cover of $\mathcal{H}$ based on the empirical $\ell_1$ norm, namely $\ell_1(f, g) = (m + u)^{-1} \sum_{i=1}^{m+u} |f(x_i) - g(x_i)|$ for any $f, g \in \mathcal{H}$. We restrict ourselves here to the binary case here.*

## 2.2 Vapnik's Implicit Bounds

Fix some hypothesis $h \in \mathcal{H}$ and suppose that $h$ makes $k_h$ errors on the full sample (i.e. $k_h = (m + u)R_h(X_{m+u})$). Consider a random choice of the training set $S_m$ from the full sample, and let $B(r, k_h, m, u)$ be the probability that $h$ makes exactly $r$ errors over the training set $S_m$. This probability is by definition the hypergeometric distribution, given by

$$B(r, k_h, m, u) \triangleq \frac{\binom{k_h}{r}\binom{m+u-k_h}{m-r}}{\binom{m+u}{m}}.$$

Since $m$ and $u$ are fixed, throughout this discussion we abbreviate $B(r, k_h) \triangleq B(r, k_h, m, u)$. Define

$$C(\varepsilon, k_h) \triangleq \mathbf{Pr}\{R_h(X_u) - R_h(X_m) > \varepsilon\}$$

$$= \mathbf{Pr}\left\{\frac{k_h - r}{u} - \frac{r}{m} > \varepsilon\right\}$$

$$= \sum_r B(r, k_h),$$

where the summation is over all values of $r$ such that $\max\{k_h - u, 0\} \leq r \leq \min\{m, k_h\}$ and

$$\frac{k_h - r}{u} - \frac{r}{m} > \varepsilon. \tag{5}$$

Define

$$\Gamma(\varepsilon) \triangleq \max_k C\left(\sqrt{\frac{k}{m+u}} \cdot \varepsilon, k\right). \tag{6}$$

We now state Vapnik's implicit bound for transduction. The bound is slightly adapted to incorporate a prior probability over $\mathcal{H}$ (the original bound deals with a uniform prior). Also, the original bound is two-sided and the following theorem is one-sided.[6]

**Theorem 8 (Vapnik 1982)** *Let $\delta$ be given, let $\mathbf{p}$ be a prior distribution over $\mathcal{H}$, and let $\varepsilon^*(h)$ be the minimal value of $\varepsilon$ that satisfies $\Gamma(\varepsilon) \le \mathbf{p}(h)\delta$. Then, with probability at least $1 - \delta$, for all $h \in \mathcal{H}$,*

$$\frac{R_h(X_u) - R_h(X_m)}{\sqrt{R_h(X_{m+u})}} < \varepsilon^*(h). \tag{7}$$

**Proof** Using the union bound we have,

$$
\begin{aligned}
Q &= \mathbf{Pr}\left\{\exists h \in \mathcal{H} \ : \ \frac{R_h(X_u) - R_h(X_m)}{\sqrt{R_h(X_{m+u})}} > \varepsilon^*(h)\right\} \\
&= \mathbf{Pr}\left\{\exists h \in \mathcal{H} \ : \ R_h(X_u) - R_h(X_m) > \sqrt{R_h(X_{m+u})}\varepsilon^*(h)\right\} \\
&= \mathbf{Pr}\left\{\exists h \in \mathcal{H} \ : \ R_h(X_u) - R_h(X_m) > \sqrt{\frac{k_h}{m+u}}\varepsilon^*(h)\right\} \\
&\le \sum_{h \in \mathcal{H}} C\left(\sqrt{\frac{k_h}{m+u}}\varepsilon^*(h), k_h\right) \\
&\le \sum_{h \in \mathcal{H}} \Gamma(\varepsilon^*(h)) \\
&\le \sum_{h \in \mathcal{H}} \mathbf{p}(h)\delta = \delta.
\end{aligned}
\tag{8}
$$

Note: The convention $\frac{R_h(X_u) - R_h(X_m)}{\sqrt{R_h(X_{m+u})}} = 0$ is used whenever $R_h(X_u) = R_h(X_m) = R_h(X_{m+u}) = 0$. ∎

It is not hard to convert the bound (7) to the "standard" form (i.e. expressed as empirical error plus some complexity term). Squaring both sides of (7) and then substituting $\frac{m}{m+u}R_h(X_m) + \frac{u}{m+u}R_h(X_u)$ for $R_h(X_{m+u})$ we get a quadratic inequality where the "unknown" variable is $R_h(X_u)$. Solving for $R_h(X_u)$ yields the following result (as in Vapnik, 1998, Equation (8.15)).

**Corollary 9** *Under the conditions of Theorem 8,*

$$R_h(X_u) \le R_h(X_m) + \frac{\varepsilon^*(h)^2 u}{2(m+u)} + \varepsilon^*(h)\sqrt{R_h(X_m) + \left(\frac{\varepsilon^*(h)u}{2(m+u)}\right)^2}.$$

**Remark 10** *Theorem 8 focuses on the relative deviation of the divergence $R_h(X_u) - R_h(X_m)$, and generates a bound that is particularly tight in cases where the empirical error $R_h(X_m)$ is very small. A very similar but simpler version of Vapnik's bound concerns the "absolute"* deviation is tighter in other cases of interest. The absolute bound is obtained as follows. Define, instead of (6),

$$\Gamma(\varepsilon) \triangleq \max_k C(\varepsilon, k),$$

and let $\varepsilon^*(h)$ be the minimal value of $\varepsilon$ that satisfies $\Gamma(\varepsilon) \leq \mathbf{p}(h)\delta$, for any given $\delta$. Then, the absolute bound states that with confidence at least $1 - \delta$,

$$R_h(X_u) < R_h(X_m) + \varepsilon^*(h). \tag{9}$$

---

6. Specifically, in the original bound (Vapnik, 1982) the two-sided condition $|\frac{k_h - r}{u} - \frac{r}{m}| > \varepsilon$ is used instead of condition (5).

This result is presented by Bottou, Cortes, and Vapnik (1994).

**Remark 11** The bound of Corollary 9, and the bound (9), are rather tight. Possible sources of slackness are only introduced through the utilization of the union bound in (8) and the definition of $\Gamma$ in (6). However, note that $\varepsilon^*(h)$ is a complicated implicit function of $m$, $u$, $\mathbf{p}(h)$ and $\delta$ leading to a bound that is difficult to interpret and (as noted also by Vapnik) must be tabulated by a computer in order to be used.

Note that a related result has recently been presented by Blum and Langford (2003). Specifically, Theorem 6 in that paper states a similar implicit bound, based on direct calculation of the hypergeometric tail. However, Vapnik's bound was originally proved for a uniform prior over the hypothesis class, and the extension to general priors was first proposed by Blum and Langford (2003).[7]

## 3. Concentration Inequalities for Sampling without Replacement

In this section we present several concentration inequalities that will be used in Section 4 to develop PAC-Bayesian bounds for transduction. As discussed in Section 2 (see Remark 1), sampling without replacement leads to dependent data, precluding direct application of standard large deviation bounds devised for independent samples. Here we present several concentration inequalities for sampling without replacement.

### 3.1 Inequalities Based on Reduction to Independence

Even though sampling without replacement leads to dependent samples, Hoeffding (1963) pointed out a simple procedure to transform the problem into one involving independent data. While this procedure leads to non-trivial bounds it involves some loss in tightness (see Section 3.2).

**Lemma 12 (Hoeffding 1963)** Let $\mathcal{C} = \{c_1, \ldots, c_N\}$ be a finite set with $N$ elements, let $\{X_1, \ldots, X_m\}$ be chosen uniformly at random **with replacement** from $\mathcal{C}$, and let $\{Z_1, \ldots, Z_m\}$ be chosen uniformly at random **without replacement** from $\mathcal{C}$.[8] Then, for any continuous and convex real-valued function $f(x)$, $\mathbf{E}f\left(\sum_{i=1}^m Z_i\right) \leq \mathbf{E}f\left(\sum_{i=1}^m X_i\right)$.

---

7. Also, the bound of Blum and Langford (2003) may be tighter than the Vapnik's bounds in some cases of interest. A careful numerical comparison should be conducted to determine the relative advantage of each of these bounds.

8. Note that the variables $\{X_1, \ldots, X_m\}$ are *independent*, while $\{Z_1, \ldots, Z_m\}$ are *dependent*.

Lemma 12 can be used in order to generate standard exponential bounds, as proposed by Hoeffding (1963). We first introduce some notation which will be used in the sequel. Let $\nu$ and $\mu$ be two real numbers in $[0, 1]$. We use the following definitions for the binary entropy and binary KL-divergence, respectively.

$$H(\nu) \triangleq -\nu \log \nu - (1 - \nu) \log(1 - \nu),$$

$$D(\nu \| \mu) \triangleq \nu \log \frac{\nu}{\mu} + (1 - \nu) \log \frac{1 - \nu}{1 - \mu}.$$

**Theorem 13 (Hoeffding, 1963)** *Let $\mathcal{C} = \{c_1, \ldots, c_N\}$ be a finite set of non-negative bounded real numbers, $c_i \leq B$, and set $\bar{c} = (1/m) \sum_{i=1}^{m} c_i$. Let $Z_1, \ldots, Z_m$, be random variables obtaining their values by sampling $\mathcal{C}$ uniformly at random **without** replacement. Set $Z = (1/m) \sum_{i=1}^{m} Z_i$. Then, for any $\varepsilon \leq 1 - \bar{c}/B$,*

$$\mathbf{Pr}\{Z - \mathbf{E}Z \geq \varepsilon\} \leq \exp\left\{-mD\left(\frac{\bar{c}}{B} + \varepsilon \,\middle\|\, \frac{\bar{c}}{B}\right)\right\} \tag{10}$$

$$\leq \exp\left\{-\frac{2m\varepsilon^2}{B^2}\right\} \tag{11}$$

*Similar bounds hold for $\mathbf{Pr}\{\mathbf{E}Z - Z \geq \varepsilon\}$.*

The key to the proof Theorem 13 is the application of Lemma 12 with $f(\sum_i Z_i) = \exp\{\sum_i (Z_i - \mathbf{E}Z_i)\}$ and the utilization of the Chernoff-Hoeffding bounding technique (see Hoeffding, 1963, for the details).

## 3.2 Sampling Without Replacement - Direct Inequalities

In this section we consider approaches which directly establish exponential bounds for sampling without replacement. As opposed to Vapnik's results (1982, 1998), which provide tight but implicit bounds, we aim at bounds which depend *explicitly* on all parameters of interest. Note that the bound of Theorem 13 does not depend on all the parameters (in particular, the population size $N$ does not appear and clearly, small population size should affect the convergence rate). One may expect that bounds developed directly for sampling without replacement should be tighter than those based on reduction to independence. The reason for this is as follows. Assume, we have sampled $k$ out of $N$ points without replacement. The next point is to be sampled from a set of $N - k$ rather than $N$ points, which would be the case in sampling with replacement (where the samples are independent). The successive reduction in the size of the sampled set reduces the 'randomness' of the newly sampled point as compared to the independent case. This intuition is at the heart of Serfling's improved bound (Serfling, 1974), which is stated next. This result holds for general bounded loss functions and is established by a careful utilization of martingale techniques combined with Chernoff's bounding method.

**Theorem 14 (Serfling, 1974)** *Let $\mathcal{C} = \{c_1, \ldots, c_N\}$ be a finite set of bounded real numbers, $|c_i| \leq B$. Let $Z_1, \ldots, Z_m$, be random variables obtaining their values by sampling $\mathcal{C}$ uniformly at random **without** replacement. Set $Z = \frac{1}{m} \sum_{i=1}^{m} Z_i$. Then,*

$$\mathbf{Pr}\{Z - \mathbf{E}Z \geq \varepsilon\} \leq \exp\left\{-\left(\frac{2m\varepsilon^2}{B^2}\right)\left(\frac{N}{N - m + 1}\right)\right\}, \tag{12}$$

and similarly for $\mathbf{Pr}\{\mathbf{E}Z_m - Z_m \geq \varepsilon\}$.

Compared to the bound of Theorem 13, the bound in (12) is always tighter than Hoeffding's second bound (11) when $N/(N - m + 1) > 1$ (i.e. when $m > 1$). When applied to our transduction setup (see Section 4.1) we take $N = m + u$ and the advantage is maximized when $(m + u)/(u + 1)$ is maximized. Thus, considering only the convergence rate obtained by sampling without replacement, one may expect that the fastest rates should be obtained when $u$ assumes the smallest possible value (e.g. $u = 1$) and not surprisingly, the advantage over the bound of Theorem 13 vanishes as $u \to \infty$.

In the case where the $c_i$ are binary variables, the bound in Theorem 14 can be improved, by using a proof based on a counting argument. The following theorem and proof is based on a simple consequence of Lemma 2.1.33 by Dembo and Zeitouni (1998).

**Theorem 15** Let $\mathcal{C} = \{c_1, \ldots, c_N\}$, $c_i \in \{0, 1\}$, be a finite set of binary numbers, and set $\bar{c} = (1/N) \sum_{i=1}^{N} c_i$. Let $Z_1, \ldots, Z_m$, be random variables obtaining their values by sampling $\mathcal{C}$ uniformly at random **without** replacement. Set $Z = (1/m) \sum_{i=1}^{m} Z_i$ and $\beta = m/N$. Then, if [9] $\varepsilon \leq \min\{1 - \bar{c}, \bar{c}(1 - \beta)/\beta\}$,

$$\mathbf{Pr}\{Z - \mathbf{E}Z > \varepsilon\} \leq \exp\left\{-mD(\bar{c} + \varepsilon \| \bar{c}) - (N - m)\, D\left(\bar{c} - \frac{\beta\varepsilon}{1 - \beta}\, \Big\|\, \bar{c}\right) + 7\log(N + 1)\right\}.$$

**Proof** Denote by $N_0$ and $N_1$ the number of appearances in $\mathcal{C}$ of $c_i = 0$ and $c_i = 1$, respectively. Let $m_0$ and $m_1$ be integers $0 \leq m_0, m_1 \leq m$, such that $m_0 + m_1 = m$. The probability of observing $m_1$ appearances of '1' (and thus $m_0$ appearances of '0' ) in a random sub-sample selected without replacement is the number of $m$-tuples resulting in $m_0(m_1)$ appearances of $0(1)$ in the subsample, divided by the overall number of $m$-tuples,

$$\mathbf{Pr}\left\{\sum_{i=1}^{m} Z_i = m_1\right\} = \frac{\binom{N_1}{m_1}\binom{N_0}{m_0}}{\binom{N}{m}}.$$

Setting $\mu = \bar{c}$, the probability that $Z = (1/m)\sum_{i=1}^{m} Z_i$ is greater than $\nu \triangleq \mu + \varepsilon = m_1/m$ (for some natural number $m_1$) is then given by

$$\mathbf{Pr}\left\{\frac{1}{m}\sum_{i=1}^{m} Z_i > \nu\right\} = \sum_{m_1 = \lceil m\nu \rceil}^{m} \frac{\binom{N_1}{m_1}\binom{N_0}{m_0}}{\binom{N}{m}}.$$

Using the Stirling bound we have that

$$\max_{1 \leq m \leq N}\left|\log\binom{N}{m} - NH\left(\frac{m}{N}\right)\right| \leq 2\log(N + 1).$$

We thus find that

$$\mathbf{Pr}\left\{\frac{1}{m}\sum_{i=1}^{m} Z_i > \nu\right\} \leq \sum_{m_1 = m\nu}^{m} \exp\left\{N_0 H\left(\frac{m_0}{N_0}\right) + N_1 H\left(\frac{m_1}{N_1}\right) - NH\left(\frac{m}{N}\right) + 6\log(N + 1)\right\}$$

$$= \sum_{m_1 = m\nu}^{m} \exp\left\{-mD(\nu \| \mu) - (N - m)D\left(\frac{\mu - \beta\nu}{1 - \beta}\, \Big\|\, \mu\right) + 6\log(N + 1)\right\}.$$

---

9. The second condition, $\varepsilon \leq \bar{c}(1 - \beta)/\beta$, simply guarantees that the number of 'ones' in the sub-sample does not exceed their number in the original sample.

The claim is concluded by upper bounding the sum by the product of the number of terms in the sum and the maximal term. It is easy to verify that the maximal summand is attained at $\nu = \mu$. Assuming that $\nu > \mu$, and using the convexity of $D(\nu\|\mu)$ with respect to $\nu$, we conclude that the largest contribution to the sum is attained at $m_1 = m\nu$, yielding the bound

$$\mathbf{Pr}\left\{\frac{1}{m}\sum_i Z_i > \nu\right\} \leq m(1-\nu)\exp\left\{-mD(\nu\|\mu) - (N-m)\ D\left(\left.\frac{\mu-\beta\nu}{1-\beta}\right\|\mu\right) + 6\log(N+1)\right\}$$

$$\leq \exp\left\{-mD(\nu\|\mu) - (N-m)D\left(\left.\frac{\mu-\beta\nu}{1-\beta}\right\|\mu\right) + 7\log(N+1)\right\},$$

which establishes the claim upon setting $\nu = \bar{c} + \varepsilon$. ∎

Note that the proof of the last bound does not rely on the Chernoff-Hoeffding bounding technique as used by Hoeffding (1963) and in many other derivations, but rather on a direct counting argument.

**Remark 16** *We are aware of another concentration inequality for sampling without replacement, which also applies to binary variables. This inequality, by Hush and Scovel (2003), is an extension of a result by Vapnik (1998, Sec. 4.13). While Vapnik's result concerns the case $m = N/2$ ($u = m$ in the transduction setup), the Hush and Scovel result considers the general case of arbitrary $m$ and $N$. The transduction bound we obtain using the Hush and Scovel concentration inequality is more complex but is qualitatively the same as the bound of Corollary 23 that we later present. We therefore omit this bound.*

## 4. PAC-Bayesian Transduction Bounds

In this section we present general error bounds for transductive learning. Our bounds can be viewed as extensions of McAllester's PAC-Bayesian inductive bounds (1999, 2003a, 2003b). In Section 4.1 we focus on simple randomized learning algorithms which are typically referred to as 'Gibbs algorithms'. Then in Section 4.2 we consider a standard deterministic setting. In the case of binary classification the bounds for deterministic learning are comparable to Vapnik's bounds presented in Section 2.2. Unlike the implicit but tight PAC-Bayesian bound of Theorem 8 (and Corollary 9), the new bounds are somewhat looser but explicit.

### 4.1 Bounds for Transductive Gibbs Learning

We present two bounds. The first is a rather immediate extension of McAllester's bound (2003b, equation (6)), using a reduction to independence, as discussed in Section 3.1. The second bound is based on the 'direct approach' and is considerably tighter in many cases of interest.

Within the original inductive setting, the selection of the prior distribution $\mathbf{p}$ in the PAC-Bayesian bounds must be made prior to observing the data. As we later show, in the present transductive setting it is possible to obtain much more compact (and effective) priors by first observing the full input sample $X_{m+u}$ and using it to construct a prior $\mathbf{p} = \mathbf{p}(X_{m+u})$. However, as shown by McAllester (2003a), under certain conditions it is possible to provide performance guarantees even if we select a "posterior" distribution over the hypothesis space *after* observing the labels of the training points $X_m$. The guarantee is provided for a *Gibbs algorithm*, which is simply a *stochastic* classifier defined as follows. Let $\mathbf{q}$ be any distribution over $\mathcal{H}$. The corresponding Gibbs classifier, denoted by $G_{\mathbf{q}}$, classifies any new instance using a randomly chosen hypothesis $h \in \mathcal{H}$, with $h \sim \mathbf{q}$ (i.e. each new instance is classified with a potentially new random classifier).

For Gibbs classifiers we now extend definition (1) as follows. Let $Z = x_1, \dots, x_{|Z|}$ be any set of samples and let $G_{\mathbf{q}}$ be a Gibbs classifier over $\mathcal{H}$. The (expected) risk of $G_{\mathbf{q}}$ over $Z$ is

$$R_{G_{\mathbf{q}}}(Z) \stackrel{\triangle}{=} \mathbf{E}_{h \sim \mathbf{q}} \left\{ \frac{1}{|Z|} \sum_{i=1}^{|Z|} \ell(h(x_i), \phi(x_i)) \right\}.$$

As before, when $Z = X_m$ (the training set) we use the standard notation $\hat{R}_{G_{\mathbf{q}}}(S_m) = R_{G_{\mathbf{q}}}(X_m)$.

The first risk bound we state for transductive Gibbs classifiers is a simple extension of the recent inductive generalization bound for Gibbs classifiers presented by McAllester (2003b). The new transductive bound relies on reduction to independence and its proof follows almost exactly the proof of the original inductive result. We therefore omit the proof but note that the inductive bound relies on the variant of Theorem 13 (inequality (10)) for sampling with replacement. The new bound is obtained by bounding the divergence between the $R_h(X_{m+u})$ and $\hat{R}_h(S_m)$ now relying on inequality (10), which concerns sampling without replacement. The bound on the transductive risk $R_h(X_u)$ is obtained using the following simple generalization of Lemma 5, stating that for all $\mathbf{q}$ and $C$,

$$R_{G_{\mathbf{q}}}(X_{m+u}) \leq \hat{R}_{G_{\mathbf{q}}}(S_m) + C \quad \Leftrightarrow \quad R_{G_{\mathbf{q}}}(X_u) \leq \hat{R}_{G_{\mathbf{q}}}(S_m) + \frac{m+u}{u} \cdot C. \quad (13)$$

**Theorem 17 (Gibbs Classifiers)** *Let* $X_{m+u} = X_m \cup X_u$ *be the full sample. Let* $\mathbf{p} = \mathbf{p}(X_{m+u})$ *be a (prior) distribution over* $\mathcal{H}$ *that may depend on the full sample. Let* $\delta \in (0, 1)$ *be given. Then with probability at least* $1 - \delta$ *over the choices of* $S_m$ *(from the full sample) the following bound holds for any distribution* $\mathbf{q}$,

$$R_{G_{\mathbf{q}}}(X_u) \leq \hat{R}_{G_{\mathbf{q}}}(S_m) + \left( \frac{m+u}{u} \right) \left( \sqrt{\frac{2\hat{R}_{G_{\mathbf{q}}}(S_m) \left( D(\mathbf{q}||\mathbf{p}) + \ln \frac{m}{\delta} \right)}{m-1}} + \frac{2 \left( D(\mathbf{q}||\mathbf{p}) + \ln \frac{m}{\delta} \right)}{m-1} \right),$$
$$(14)$$

*where* $D(\cdot||\cdot)$ *is the familiar Kullback-Leibler (KL) divergence (see e.g. Cover & Thomas, 1991).*

Notice that when $\hat{R}_{G_{\mathbf{q}}}(S_m) = 0$ (the so-called "realizable case") fast convergence rates of order $1/m$ are possible when $u$ is sufficiently large (i.e. $u = \omega(m)$).

The next risk bound we present for transductive Gibbs *binary* classifiers relies on the "direct" concentration inequality of Theorem 15, for sampling without replacement. The proof is based on the proof technique recently presented by McAllester (2003b), which, in turn, is based on the results of Langford and Shawe-Taylor (2002) and Seeger (2003).

**Theorem 18 (Binary Gibbs Classifiers)** *Let the conditions of Theorem 17 hold, and assume the loss is binary. Then with probability at least $1 - \delta$ over the choices of $S_m$ (from the full sample) the following bound holds for any distribution* $\mathbf{q}$,

$$R_{G_\mathbf{q}}(X_u) \leq \hat{R}_{G_\mathbf{q}}(S_m) + \sqrt{\left(\frac{2\hat{R}_{G_\mathbf{q}}(S_m)(m+u)}{u}\right)\frac{D(\mathbf{q}\|\mathbf{p}) + \ln\frac{m}{\delta} + 7\log(m+u+1)}{m-1}}$$

$$+ \frac{2\left(D(\mathbf{q}\|\mathbf{p}) + \ln\frac{m}{\delta} + 7\log(m+u+1)\right)}{m-1}$$

Before we prove Theorem 18 observe that when $\hat{R}_{G_\mathbf{q}}(S_m) = 0$ (the "realizable case") the bound converges even if $u = 1$. In contrast, the bound of Theorem 17 diverges in the realizable case for any $u = O(\sqrt{m})$.

For proving Theorem 18 we quote without proof two results by McAllester (2003b).

**Lemma 19 (Lemma 5, McAllester, 2003b)** *Let $X$ be a random variable satisfying* $\mathbf{Pr}\{X > x\} \leq e^{-mf(x)}$ *where $f(x)$ is non-negative. Then* $\mathbf{E}\left[e^{(m-1)f(X)}\right] \leq m$.

**Lemma 20 (Lemma 8, McAllester, 2003b)** $\mathbf{E}_{x \sim \mathbf{q}}[f(x)] \leq D(\mathbf{q}\|\mathbf{p}) + \ln \mathbf{E}_{x \sim \mathbf{p}} e^{f(x)}$.

**Proof of Theorem 18:** The proof is based on the ideas by McAllester (2003b). Define

$$\hat{\nu}_h \triangleq \hat{R}_h(S_m) \quad ; \quad \mu_h \triangleq R_h(X_{m+u}).$$

Let

$$f_h(\nu) \triangleq D(\nu\|\mu_h) + \frac{u}{m}D\left(\frac{\mu_h - \beta\nu}{1 - \beta}\|\mu_h\right) - \frac{7}{m}\log(m+u+1).$$

¿From Lemma 20 we have that

$$\mathbf{E}_{h \sim \mathbf{q}}[(m-1)f_h(\nu)] \leq D(\mathbf{q}\|\mathbf{p}) + \ln \mathbf{E}_{h \sim \mathbf{p}}\left[e^{(m-1)f_h(\nu)}\right]. \tag{15}$$

An upper bound on $\mathbf{E}_{h \sim \mathbf{p}}\left[e^{(m-1)f_h(\nu)}\right]$ may be obtained by the following argument. From Theorem 15 we have that

$$\mathbf{Pr}\{\hat{\nu}_h > \nu\} \leq \exp\{-mf_h(\nu)\}.$$

Lemma 19 then implies that for any $h$

$$\mathbf{E}_{\Sigma_m}\left[e^{(m-1)f_h(\hat{\nu}_h)}\right] \leq m,$$

which implies that

$$\mathbf{E}_{\Sigma_m}\mathbf{E}_{h \sim \mathbf{p}}\left[e^{(m-1)f_h(\hat{\nu}_h)}\right] \leq m,$$

from which we infer that with probability at least $1 - \delta$,

$$\mathbf{E}_{h \sim \mathbf{p}}\left[e^{(m-1)f_h(\hat{\nu}_h)}\right] \leq \frac{m}{\delta}$$

by using Markov's inequality. Substituting in (15) we find that with probability at least $1 - \delta$,

$$\mathbf{E}_{h \sim \mathbf{q}}\left[(m-1)f_h(\hat{\nu}_h)\right] \leq D(\mathbf{q}\|\mathbf{p}) + \frac{m}{\delta} . \tag{16}$$

Substituting for $f_h(\hat{\nu}_h)$, and using the convexity of the function $x \log x$, we find that

$$D(\hat{R}_{G_{\mathbf{q}}}(S_m)\|R_{G_{\mathbf{q}}}(X_{m+u})) + \frac{u}{m}D\left(\frac{R_{G_{\mathbf{q}}}(X_{m+u}) - \beta\hat{R}_{G_{\mathbf{q}}}(S_m)}{1-\beta}\|R_{G_{\mathbf{q}}}(X_{m+u})\right) - \frac{7}{m}\log(m+u+1)$$

$$\leq \frac{D(\mathbf{q}\|\mathbf{p}) + \ln\frac{m}{\delta}}{m-1} . \tag{17}$$

In order to obtain an explicit bound, we use the inequality

$$D(\nu\|\mu) \geq \frac{(\nu-\mu)^2}{2\mu},$$

and substitute this in (17) obtaining

$$\underbrace{R_{G_{\mathbf{q}}}(X_{m+u})}_{z} \leq \underbrace{\hat{R}_{G_{\mathbf{q}}}(S_m)}_{a} + \sqrt{\underbrace{R_{G_{\mathbf{q}}}(X_{m+u})}_{z}\underbrace{\left(\frac{2u}{m+u}\right)\frac{D(\mathbf{q}\|\mathbf{p}) + \ln\frac{m}{\delta} + 7\log(m+u+1)}{m-1}}_{b}}.$$

Thus we have (with probability at least $1-\delta$), $z \leq a + \sqrt{zb}$ (where $z = R_{G_{\mathbf{q}}}(X_{m+u})$). Solving for $z$ we get $z \leq a + b + \sqrt{ab}$, and using Lemma 5 yields the desired result. $\qquad\square$

**Remark 21** *It is interesting to compare the bound of Theorem 17, based on the reduction to independence approach, and that of Theorem 18 which is based on a direct concentration inequality for sampling without replacement. The complexity term in Theorem 17 is multiplied by $(m+u)/u$, while the corresponding term in Theorem 18 is multiplied by $\sqrt{(m+u)/u}$. This clearly displays the advantage of using the direct concentration bound, even though it does* not *lead to improved convergence rates in general. More importantly, for the realizable case, $\hat{R}_{G_{\mathbf{q}}}(S_m) = 0$, the bound of Theorem 18 converges to zero even for $u = 1$. This is not the case for the bound of Theorem 17.*

## 4.2 Bounds for Deterministic Learning Algorithms

In this section we present three transductive PAC-Bayesian error bounds for deterministic learning algorithms. Note that the two bounds we present for the (stochastic) Gibbs algorithms in the previous subsection can be specialized to deterministic algorithms. This is done by choosing a "posterior" $\mathbf{q}$ which assigns probability 1 to one desired hypothesis $h \in \mathcal{H}$. In doing so the term $D(\mathbf{q}\|\mathbf{p})$ reduces to $\log(1/\mathbf{p}(h))$. For example, the bound of Theorem 17 reduces to

$$R_h(X_u) \leq \hat{R}_h(S_m) + \left(\frac{m+u}{u}\right)\left(\sqrt{\frac{2\hat{R}_h(S_m)\left(\log\frac{1}{\mathbf{p}(h)} + \ln\frac{m}{\delta}\right)}{m-1}} + \frac{2\left(\log(\frac{1}{\mathbf{p}(h)} + \ln\frac{m}{\delta}\right)}{m-1}\right), \tag{18}$$

which applies to any bounded loss function.

The following bound relies of the Serfling concentration inequality presented in Theorem 14 and applies to any bounded loss function.[10]

**Theorem 22** *Let $X_{m+u} = X_m \cup X_u$ be the full sample and let $\mathbf{p} = \mathbf{p}(X_{m+u})$ be a (prior) distribution over $\mathcal{H}$ that may depend on the full sample. Assume that $\ell(h(x), y) \in [0, B]$ and let $\delta \in (0, 1)$ be given. Then, with probability at least $1 - \delta$ over choices of $S_m$ (from the full sample) the following bound holds for any $h \in \mathcal{H}$,*

$$R_h(X_u) \le \hat{R}_h(S_m) + B\sqrt{\left(\frac{m+u}{u}\right)\left(\frac{u+1}{u}\right)\left(\frac{\ln\frac{1}{\mathbf{p}(h)} + \ln\frac{1}{\delta}}{2m}\right)}. \tag{19}$$

**Proof** In our transduction setting the set $X_m$ (and therefore $S_m$) is obtained by sampling the full sample $X_{m+u}$ uniformly at random without replacement. It is not hard to see that $\mathbf{E}_{\Sigma_m}\hat{R}_h(S_m) = R_h(X_{m+u})$. Specifically,

$$\mathbf{E}_{\Sigma_m}\hat{R}_h(S_m) = \frac{1}{\binom{m+u}{m}}\sum_{S_m}\hat{R}_h(S_m) = \frac{1}{\binom{m+u}{m}}\sum_{X_m \subseteq X_{m+n}}\frac{1}{m}\sum_{x \in S_m}\ell(h(x), \phi(x)). \tag{20}$$

By symmetry, all points $x \in X_{m+u}$ are counted on the right-hand side an equal number of times; this number is precisely $\binom{m+u}{m} - \binom{m+u-1}{m} = \binom{m+u-1}{m-1}$. The result is obtained by considering the definition of $R_h(X_{m+u})$ and noting that $\binom{m+u-1}{m-1}/\binom{m+u}{m} = \frac{m}{m+u}$. Using the fact that our loss function is bounded in $[0, B]$ we apply Theorem 14 (for a fixed $h$ and $N = m + u$),

$$\mathbf{Pr}_{\Sigma_m}\left\{\mathbf{E}\hat{R}_h(S_m) - \hat{R}_h(S_m) > \varepsilon\right\} \le e^{-\frac{2m\varepsilon^2}{B^2}\left(\frac{m+u}{u+1}\right)}. \tag{21}$$

Setting $\varepsilon(h) = B\sqrt{\frac{(u+1)(\ln\frac{1}{\mathbf{p}(h)} + \ln\frac{1}{\delta})}{(m+u)2m}}$ and using the union bound we find

$$\mathbf{Pr}_{\Sigma_m}\left\{\exists h \in \mathcal{H} \text{ s.t. } R_h(X_{m+u}) - \hat{R}_h(S_m) > \varepsilon(h)\right\} \le \sum_h \exp\left\{-\frac{2m\varepsilon(h)^2}{B^2}\left(\frac{m+u}{u+1}\right)\right\}$$
$$= \sum_h \mathbf{p}(h)\delta$$
$$= \delta.$$

We thus obtain that

$$R_h(X_{m+u}) \le \hat{R}_h(S_m) + B\sqrt{\left(\frac{u+1}{m+u}\right)\left(\frac{\ln\frac{1}{\mathbf{p}(h)} + \ln\frac{1}{\delta}}{2m}\right)}. \tag{22}$$

The proof is then completed using Lemma 5. ∎

For classification using the 0/1 loss function we present one bound, which is a specialization of Theorem 18.

---

10. Although we have not utilized the Serfling inequality for devising a bound for the Gibbs algorithm, it can be done as well.

**Corollary 23** *Let the conditions of Theorem 22 hold and assume the loss is binary. Then with probability at least $1 - \delta$ over the choices of $S_m$ (from the full sample) the following bound holds for any distribution $\mathbf{q}$,*

$$R_h(X_u) \leq \hat{R}_h(S_m) + \sqrt{\left(\frac{2\hat{R}_h(S_m)(m+u)}{u}\right)\frac{\log\frac{1}{\mathbf{p}(h)} + \ln\frac{m}{\delta} + 7\log(m+u+1)}{m-1}}$$

$$+ \frac{2\left(\log\frac{1}{\mathbf{p}(h)} + \ln\frac{m}{\delta} + 7\log(m+u+1)\right)}{m-1} \quad .$$

Figures 1 and 2 compare the two bounds presented in this section with Vapnik's bound of Corollary 9. Throughout the discussion here the bound of Theorem 22 is referred to as the "Serfling bound". Figure 1 focuses on the realizable case (i.e. empirical error = 0). According to the statements of Theorem 22 and Corollary 23, the Serfling bound has a significantly slower rate of convergence in the realizable case. However, the constants (and logarithmic terms) are larger in the bound of Corollary 23. Panels (a) and (b) in Figure 1 indicate that the Serfling bound is significantly better than the bound of Corollary 23 when $u = \Omega(m)$ for the range of $m$ we consider. However, even in these cases, we know that the bound of Corollary 23 will eventually outperform the Serfling bound. We also see that the Serfling bound tracks Vapnik's bound quite well when $u = \Omega(m)$. On the other hand, Panels (c) and (d) indicate that the bound of Corollary 23 is significantly better than the Serfling bound when $u = o(m)$. The examples given are $u = \sqrt{m}$ in Panel (c) and $u = 10$ in Panel (d). Figure 2 shows these bounds for the case $\hat{R}_h(S_m) = 0.2$. Here again the Serfling bound nicely tracks Vapnik's bound and we see that the bound of Corollary 23 converges much more slowly. All the curves in Figures 1 and 2 consider the case $\mathbf{p}(h) = 1$. This assignment of the prior eliminates the influence of the union bound that is used to derive these bounds. In Figure 3 we show, for both the Vapnik and Serfling bounds, the complexity term as a function of the prior $\mathbf{p}(h)$, with $0.01 \leq \mathbf{p}(h) \leq 1$. Note that such prior assignments are realistic in the case of the transduction algorithm based on clustering that is introduced in Section 5.2.

While these plots indicate that our bounds approximate Vapnik's bound quite well in many cases of interest, we also see that for small values of $m$ (similar to those considered in the plots) one will gain in applications by using the implicit but tighter Vapnik bound (or the Blum and Langford,2003, bound).

## 5. Bounds for Specific Algorithms

PAC-Bayesian error bounds (both inductive and transductive) are interesting because they provide a very simple yet general formulation of learning. However, in order to provide more concrete statements (e.g. about specific learning algorithms) one must apply such bounds with some concrete priors (and posteriors, in the case of Gibbs learning, see Theorems 17 and 18). In the context of inductive learning, a major obstacle in deriving effective bounds[11] using the PAC-Bayesian framework is the construction of "compact priors". For example,

---

11. Informally, we say that a bound is "effective" if its complexity term vanishes with $m$ (the size of the training sample) and it is sufficiently small for "reasonable" values $m$.

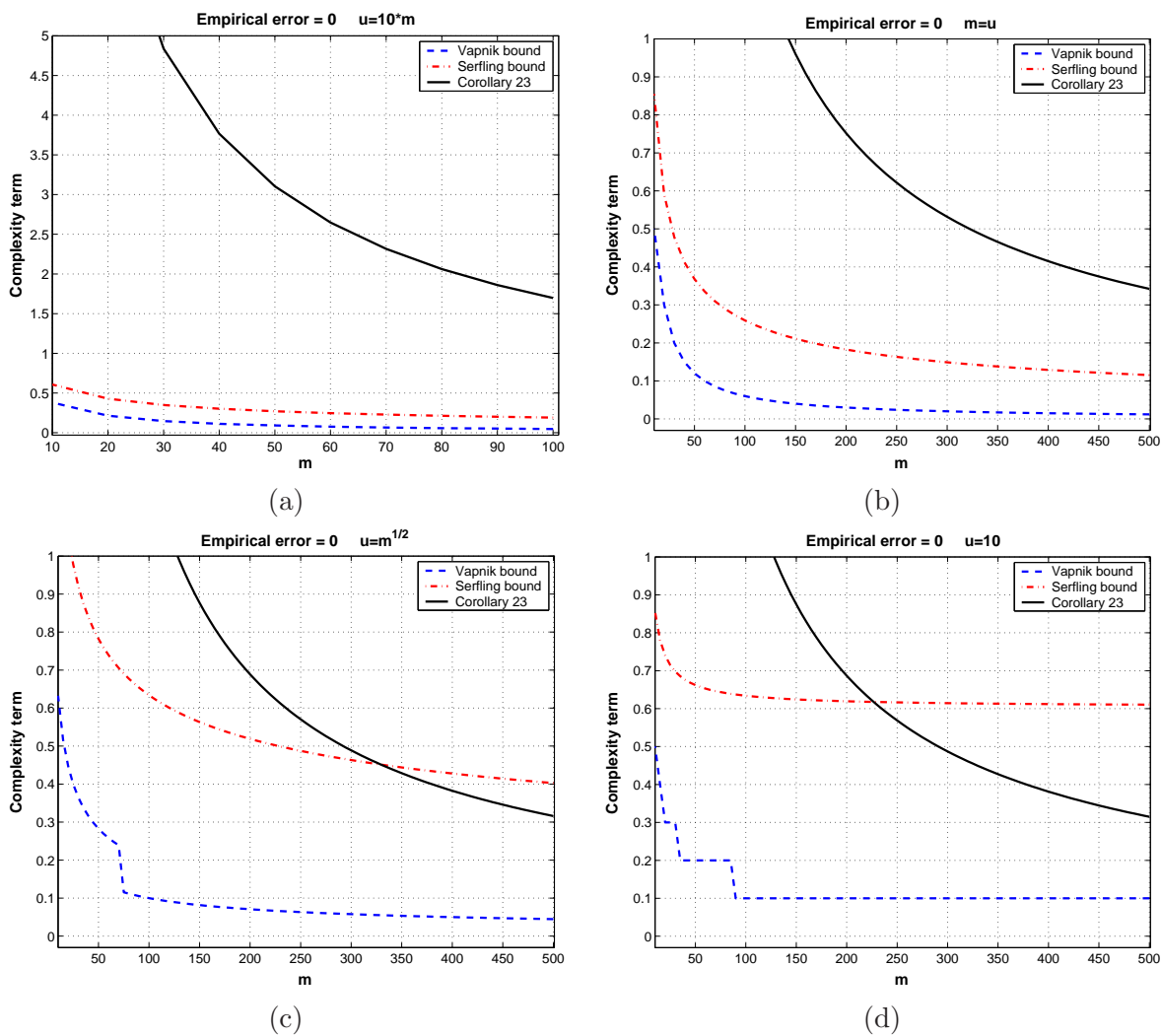

Figure 1: A comparison of Vapnik's bound (Corollary 9), the bound of Theorem 22 (denoted here by "Serfling bound") and the bound of Corollary 23. All bounds assume zero empirical error, $\delta = 0.01$ and $\mathbf{p}(h) = 1$. (a) $u = 10m$; (b) $u = m$; (c) $u = \sqrt{m}$; and (d) $u = 10$.

McAllester's generalization bound (McAllester, 1999) contains a complexity term which includes a component of the form $\ln(1/\mathbf{p}(h))$ where $\mathbf{p}$ is a prior over $\mathcal{H}$ (as in Theorem 22 and Corollary 23). The more sophisticated inductive bounds for Gibbs classifiers (McAllester, 2003a, 2003b) include a Kullback-Leibler (KL) divergence complexity component $D(\mathbf{q}\|\mathbf{p})$, where $\mathbf{p}$ is a prior over $\mathcal{H}$ and $\mathbf{q}$ is a posterior over $\mathcal{H}$ (as in Theorems 17 and 18). However, many hypothesis classes of interest are very large and even uncountably infinite. Therefore, despite the fact that these bounds apply in principle to very large $\mathcal{H}$, in a straightforward application of these PAC-Bayesian bounds, when choosing priors with a very large support

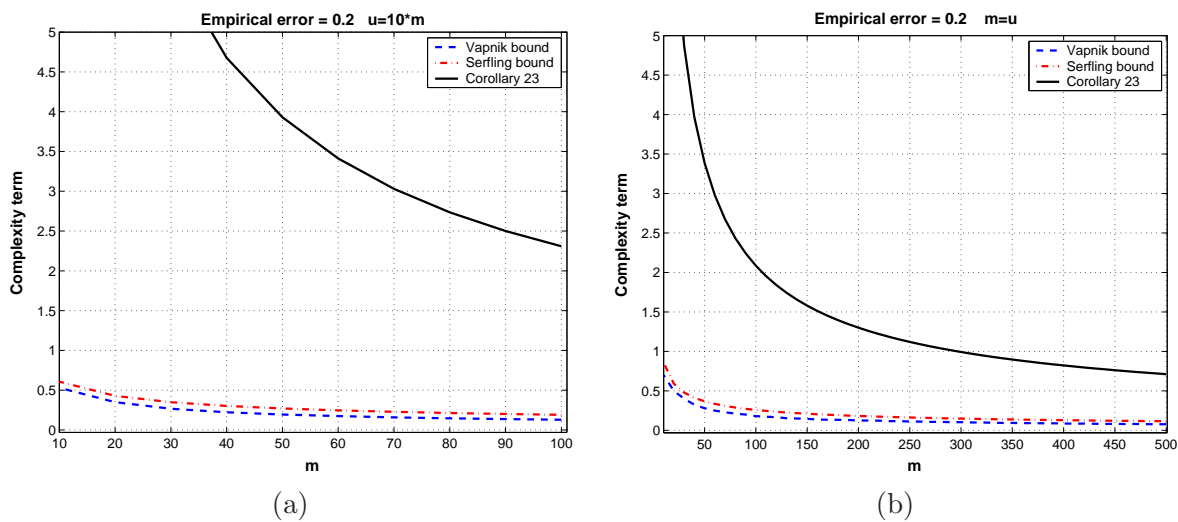

(a)                                                        (b)

Figure 2: A comparison of Vapnik's bound (Corollary 9), the bound of Theorem 22 (denoted here by "Serfling bound") and the bound of Corollary 23. All bounds assume empirical error of 0.2, $\delta = 0.01$ and $\mathbf{p}(h) = 1$. (a) $u = 10m$; (b) $u = m$.

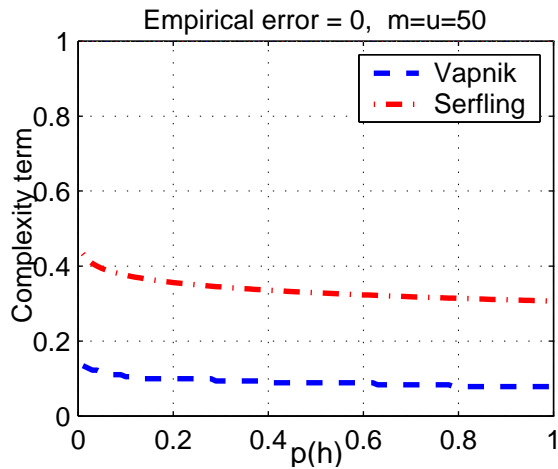

Figure 3: The complexity term of the Vapnik and Serfling bounds as a function of $\mathbf{p}(h)$ with $0.01 \leq \mathbf{p}(h) \leq 1$, $\delta = 0.01$ and $m = u = 50$.

(and possibly a posterior with a small support), the complexity terms in these bounds can diverge or at least be too large to form effective generalization bounds.[12]

---

12. Saying that we should also note that sophisticated prior choices within the inductive PAC-Bayesian framework can also lead to state-of-the-art bounds (McAllester, 2003b).

In contrast, the transductive framework provides a very convenient setting for applying PAC-Bayesian bounds. Here priors can be chosen after observing and analyzing the full sample. As already mentioned, in the case of the 0/1-loss, even if we consider a very large hypothesis space $\mathcal{H}$, after observing the full sample, the effective size of equivalence classes of hypotheses in $\mathcal{H}$ is always finite and not larger than the number of dichotomies of the full sample (see also Remark 7).

In this section we present bounds for specific learning algorithms. The first class of algorithms we consider in Section 5.1 are "compression schemes". The second class, are algorithms based on clustering. In both cases we show how to form effective priors by considering the structure of the full sample.

## 5.1 Bounds for Compression Algorithms

We propose a technique for selecting a prior $\mathbf{p}(h)$ over $\mathcal{H}$, based on the full (unlabeled) sample $X_{m+u}$. Given $m$, the learner constructs $m$ "sub-priors" $\mathbf{p}_\tau$, $\tau = 1, 2, \ldots, m$, based on the full sample $X_{m+u}$, and for the final prior, takes a uniform mixture of all these "sub-priors".

This technique generates transductive error bounds for "compression" algorithms. Let $\mathcal{A}$ be a learning algorithm. Intuitively, $\mathcal{A}$ is a "compression scheme" if it outputs the same hypothesis using a subset of the (labeled) training data.

**Definition 24** *A learning algorithm $\mathcal{A}$ (viewed as a function from samples to some hypothesis class) is a compression scheme with respect to a sample $Z$ if there is a sub-sample $Z'$, $|Z'| < |Z|$, such that $\mathcal{A}(Z') = \mathcal{A}(Z)$.*

Observe that the Support Vector Machine (SVM) approach is a compression scheme, where the set $Z'$ is determined by the set of support vectors.

Let $\mathcal{A}$ be a deterministic compression scheme and consider the full sample $X_{m+u}$. For each integer $\tau = 1, \ldots, m$, consider all subsets of $X_{m+u}$ of size $\tau$, and for each subset construct all possible dichotomies of that subset (note that we are not proposing this approach as a useful algorithm, but rather as a means to derive bounds; in practice one need not construct all these dichotomies). A deterministic algorithm $\mathcal{A}$ generates at most one hypothesis $h \in \mathcal{H}$ for each dichotomy.[13] For each $\tau$, let the set of hypotheses generated by this procedure be denoted by $\mathcal{H}_\tau$. For the rest of this discussion we assume the worst case where $|\mathcal{H}_\tau| = 2^\tau \binom{m+u}{\tau}$ (i.e. if $\mathcal{H}_\tau$ does not contains one hypothesis for each dichotomy the bounds we propose below improve). The "sub-prior" $\mathbf{p}_\tau$ is then defined to be a uniform distribution over $\mathcal{H}_\tau$.

In this way we have $m$ "sub-priors", $\mathbf{p}_1, \ldots, \mathbf{p}_m$, which are constructed using only $X_{m+u}$ (and are independent of the labels of the training set $Y_m$; also note that this construction takes place before choosing the subset $X_m$). Any hypothesis selected by the learning algorithm $\mathcal{A}$ based on the labeled sample $S_m$ and on the test set $X_u$ belongs to $\cup_{\tau=1}^m \mathcal{H}_\tau$. The motivation for this construction is as follows. Each $\tau$ can be viewed as our "guess" for the maximal number of compression points that will be utilized by a resulting classifier. For each such $\tau$ the distribution $\mathbf{p}_\tau$ is constructed over all possible classifiers that use $\tau$ compression points. By systematically considering all possible dichotomies of $\tau$ points we can characterize a relatively small subset of $\mathcal{H}$ without observing labels of the training points. Thus, each "sub-prior" $\mathbf{p}_\tau$ represents one such guess. The final prior is

$$\mathbf{p}(h) = \frac{1}{m} \sum_{\tau=1}^{m} \mathbf{p}_\tau(h). \tag{23}$$

---

13. It might be that for some dichotomies the learning algorithm will fail to construct a classifier. For example, a linear SVM in feature space without "soft margin" will fail to classify non linearly-separable dichotomies of $X_{m+u}$.

The following corollary is obtained by an application of Theorem 22 using the prior $\mathbf{p}(h)$ in (23). This results characterizes an upper bound on the divergence in terms of the observed size of the compression set of the final classifier.

**Corollary 25 (Transductive Compression Bound)** *Let the conditions of Theorem 22 hold. Let $\mathcal{A}$ be a deterministic learning algorithm leading to a hypothesis $h \in \mathcal{H}$ based on a compression set of size $s$. Then, with probability at least $1 - \delta$,*

$$R_h(X_u) \leq \hat{R}_h(S_m) + \sqrt{\left(\frac{m+u}{u}\right)\left(\frac{u+1}{u}\right)\left(\frac{s \ln\left(\frac{2e(m+u)}{s}\right) + \ln(m/\delta))}{m}\right)}. \tag{24}$$

**Proof** Recall that $\mathcal{H}_s \subseteq \mathcal{H}$ is the support set of $\mathbf{p}_s$ and that $\mathbf{p}_s(h) = 1/|\mathcal{H}_s|$ for all $h \in \mathcal{H}_s$, implying that $\ln(1/\mathbf{p}_s(h)) = |\mathcal{H}_s|$. Using the inequality $\binom{m+u}{s} \leq (e(m+u)/s)^s$ we have that $|\mathcal{H}_s| = 2^s \binom{m+u}{s} \leq (2e(m+u)/s)^s$. Using the prior (23) and substituting $\ln(m/\mathbf{p}_s(h))$ in Theorem 22 leads to the desired result. ∎

**Remark 26** *We can use Corollary 23 to get a similar result, which is sometimes tighter. Also note that compression bounds can be easily stated and proved for Gibbs learning.*

The bound (24) can be easily computed once the classifier is trained. If the size of the compression set happens to be small, we obtain a tight bound. We note that these bounds are applicable to the transductive SVM algorithms discussed by Vapnik (1998), Bennett and Demiriz (1998) and Joachims (1999). However, our bounds motivate a different strategy than the one that drives these algorithms; namely, reduce the number of support vectors! (rather than enlarge the margin, as attempted by those algorithms).

Observe the conceptual similarity of our bound to Vapnik's bound for consistent SVMs (Vapnik, 1995, Theorem 5.2), which bounds the generalization error of an SVM by the ratio between the average number of support vectors and the sample size $m$. However, Vapnik's bound can only be estimated while this bound is truly data dependent. Finally, it is interesting to compare our result to a recent inductive bound for compression schemes. In this context, Graepel *et al.* (see Theorem 5.18 in Herbrich, 2002) have derived a bound of the form

$$R(\mathsf{SVM}) \leq \frac{m}{m-s} \hat{R}(\mathsf{SVM}) + \sqrt{\frac{s \log(2em/s) + \ln(1/\delta) + 2\ln m}{2(m-s)}}, \tag{25}$$

where $R(\mathsf{SVM})$ and $\hat{R}(\mathsf{SVM})$ denote the true and empirical errors, respectively, and $s$ is the number of observed support vectors (over the training set).

## 5.2 Transductive Learning via Clustering

Some learning problems do not allow for high compression rates using compression schemes such as SVMs (i.e. the number of support vectors can sometimes be very large, see e.g. Baram et al., 2004, Table 1). A considerably stronger type of compression can often be achieved by clustering algorithms. While there is lack of formal links between entirely unsupervised clustering and classification, within a transductive setting we can provide a principled approach to using clustering algorithms for classification.

In particular, we propose the following approach: The learner applies a clustering algorithm (or a number of clustering algorithms) over the unlabeled data to generate several (unsupervised) models. The learner then utilizes the labeled data to guess labels for entire clusters (so that all points in the same cluster have the same label). In this way, a number of hypotheses are generated, one of which is then selected based on a PAC-Bayesian error bound for transduction, applied with an appropriate prior.

The natural idea of first clustering the unlabeled data and then assigning labels to clusters has been around for a long time and there are plenty of heuristic procedures that attempt to learn using this approach within a semi-supervised or transductive settings (see, e.g., Seeger, 2002, Sec. 2.1). However, to the best of our knowledge, none of the existing procedures was theoretically justified in terms of provable reduction of the true risk. In contrast, the clustering-based transduction method we propose here relies on a solid theoretical ground.

Let $\mathcal{A}$ be any (deterministic) clustering algorithm which, given the full sample $X_{m+u}$, can cluster this sample into any desired number of clusters. We use $\mathcal{A}$ to cluster $X_{m+u}$ into $1, \ldots, c$ clusters where $c \leq m$. Thus, the algorithm generates a collection of partitions of $X_{m+u}$ into $\tau = 1, 2, \ldots, c$ clusters, where each partition is denoted by $C_\tau$. For each value of $\tau$, let $\mathcal{H}_\tau$ consist of those hypotheses which assign an identical label to all points in the same cluster of partition $C_\tau$, and define the "sub-prior" $\mathbf{p}_\tau(h) = 1/2^\tau$ for each $h \in \mathcal{H}_\tau$ and zero otherwise (note that there are $2^\tau$ possible dichotomies). The final prior is a uniform mixture of all these sub-priors, $\mathbf{p}(h) = \frac{1}{c} \sum_\tau \mathbf{p}_\tau(h)$.

The learning algorithm selects a hypothesis as follows. Upon observing the labeled sample $S_m = (X_m, Y_m)$, for each of the clusterings $C_1, \ldots, C_c$ constructed above, it assigns a label to each cluster based on the majority vote from the labels $Y_m$ of points falling within the cluster (in case of ties, or if no points from $X_m$ belong to the cluster, choose a label arbitrarily). Doing this leads to $c$ classifiers $h_\tau$, $\tau = 1, \ldots, c$. We now apply Theorem 22 with the above prior $\mathbf{p}(h)$ and can choose the classifier (equivalently, number of clusters) for which the best bound holds. We thus have the following corollary of Theorem 22.

**Corollary 27** *Let $\mathcal{A}$ be any clustering algorithm and let $h_\tau$, $\tau = 1, \ldots, c$ be classifications of the test set $X_u$ as determined by clustering of the full sample $X_{m+u}$ (into $\tau$ clusters) and the training set $S_m$, as described above. Let $\delta \in (0, 1)$ be given. Then, with probability at least $1 - \delta$ over choices of $S_m$ the following bound holds for all $\tau$,*

$$R_{h_\tau}(X_u) \leq \hat{R}_{h_\tau}(S_m) + \sqrt{\left(\frac{m+u}{u}\right)\left(\frac{u+1}{u}\right)\left(\frac{\tau + \ln \frac{c}{\delta}}{2m}\right)} \qquad (26)$$

**Remark:** Note that when $m = u$ we get the bound

$$R_{h_\tau}(X_u) \leq \hat{R}_{h_\tau}(S_m) + \sqrt{\frac{\left(1 + \frac{1}{m}\right)\left(\tau + \ln \frac{c}{\delta}\right)}{m}}.$$

Also, in the case of the 0/1 loss we can use Corollary 23 to obtain significantly faster rates in the realizable case or when the training error is very small. We see that these bounds can be rather tight when the clustering algorithm is successful (e.g. when it captures the class structure in the data using a small number of clusters). Note however, that in practice one can significantly benefit by the faster rates that can be achieved utilizing Vapnik's implicit bounds presented in Section 2.2 (or the bound of Blum and Langford, 2003). Clearly, any PAC-Bayesian bound for transduction can be plugged-in within this scheme and tighter bounds should yield better performance.

Corollary 27 can be extended in a number of ways. One simple extension is the use of an *ensemble* of clustering algorithms. Specifically, we can concurrently apply $k$ different clustering algorithms (using each algorithm to cluster the data into $\tau = 1, \ldots, c$ clusters). We thus obtain $kc$ hypotheses (partitions of $X_{m+u}$). By a simple application of the union bound we can replace $\ln \frac{c}{\delta}$ by $\ln \frac{kc}{\delta}$ in Corollary 27 and guarantee that $k$ bounds hold simultaneously for all the $k$ clustering algorithms (with probability at least $1 - \delta$). We thus choose the hypothesis which minimizes the resulting bound.[14] This extension is particularly attractive since typically without prior knowledge we do not know which clustering algorithm will be effective for the dataset at hand.

To conclude this section we note that El-Yaniv and Gerzon (2004) recently presented empirical studies of the above clustering approach. These empirical evaluations on a variety of real world datasets demonstrate the effectiveness of the proposed approach.

## 6. Concluding Remarks

We presented general explicit PAC-Bayesian bounds for transductive learning. We also developed a new prior construction technique which effectively derives tight data-dependent error bounds for compression schemes and for transductive learning algorithms based on clustering.

With the exception of Theorem 22, which holds for any bounded loss function, all our transductive error bounds were presented within the simplest binary classification setting (i.e. with the 0/1-loss function and with non-stochastic labels). However, these results can be easily extended to multi-class problems and to stochastic labels. We hope that these bounds and the prior construction technique will be useful as a starting point for deriving error bounds for other known algorithms and for developing new types of transductive learning algorithms.

We emphasize, however, that in the case of classification (i.e. the 0/1 loss), implicit but tighter error bounds for transduction were already known (e.g. Vapnik's result as stated in Corollary 9). Our bounds are explicit and can therefore be useful for interpreting and characterizing the functional dependency on the problem parameters.

---

14. A better approach to combine the $k$ clustering algorithms, especially if we expect that some of the algorithms will generate identical clusterings, is to construct one "big" prior for all of them.

In applications, when using compression schemes or our clustering-based transduction approach, one can plug-in any other PAC-Bayesian transductive bound (implicit or explicit). For example, one can benefit by using the tighter Vapnik bound. In (El-Yaniv & Gerzon, 2004) and (Banerjee & Langford, 2004), the authors present empirical studies of the clustering approach of Section 5.2 by plugging in the Vapnik implicit bound and the similar implicit bound from (Blum & Langford, 2003), respectively. These empirical studies indicate that the proposed clustering-based transductive scheme can lead to state-of-the-art algorithms.

An interesting feature of any transduction error bound for Setting 1, is that it holds for "individual samples"; that is, the full sample $X_{m+u}$ need not be sampled i.i.d. from a distribution and moreover, in this setting one cannot assume that it is sampled from a fixed distribution at all! Therefore, results for this setting must hold for any given sample. In this sense, the transductive bounds within Setting 1 are considerably more robust than standard bounds in the inductive setting.

We conclude with some open questions and research directions.

1. All our results are obtained within Vapnik's "Setting 1" of transduction, which must consider any arbitrary choice of the full sample. Considering Theorem 2 and Remark 4, it would be interesting to see if tighter results are possible within the probabilistic Setting 2.

2. An interesting direction for future research could be the construction of more sophisticated priors. For example, in our compression bound (Corollary 25), for each number $s$ of compression points we assigned the same prior to each dichotomy of every $s$-subset. However, in practice, when there is structure in the data, the vast majority of all these subsets and dichotomies cannot "explain" the data and should not be assigned a large prior.

3. The bounds derived in this paper are based on a contribution from the deviation of a single hypothesis and a utilization of the union bound in the PAC-Bayesian style. More refined approaches, for example those based on McDiarmid's inequality (McDiarmid, 1989) and the entropy method (Boucheron, Lugosi, & Massart, 2003), are able to eliminate the union bound altogether. It would be interesting to see if such approaches can lead to tighter bounds in the current setting.

4. When considering arbitrary (bounded) loss functions, the basic observation for transduction (due to Vapnik) that the effective cardinality of the hypothesis class (in the case of classification) is finite, is not necessarily valid. However, it is likely that one can still benefit from the availability of the full sample. One possible approach, mentioned in Remark 7, would be to construct an empirical $\epsilon$-cover of $\mathcal{H}$ based on the $\ell_1$ norm.

5. Finally, we note that the major challenge, of determining a precise relation between the inductive and transductive learning schemes remains open. Of particular interest would be to determine the relation between the inductive *semi-supervised* setting (where the learner is also given a set of unlabeled points, but is required to induce an hypothesis for the entire space) and transduction. Our bounds suggest that transduction does not allow for learning rates that are faster than induction (as a function of $m$). On the other hand, it appears that the bounds we obtain for clustering-based transduction can be tighter than any known bound for a specific inductive learning effective algorithm.

## Acknowledgments

The work of Ran El-Yaniv and Ron Meir was partially supported by the Technion V.P.R. fund for the promotion of sponsored research and the partial support of the PASCAL network of excellence. Support from the Ollendorff center of the department of Electrical Engineering at the Technion is also acknowledged. We also thank anonymous referees and Dmitry Pechony for their useful comments.

## Appendix A. Proof of Theorem 2

**Proof** The proof we present is identical to Vapnik's original proof and is provided for the sake of self-completeness. Let $\mathcal{A}$ be some learning algorithm choosing an hypothesis $h_{\mathcal{A}} \in \mathcal{H}$ based on $S_m \cup X_u$. Define

$$
\begin{aligned}
C_{\mathcal{A}}(x_1, y_1; \ldots; x_{m+u}, y_{m+u}) &= \left| \frac{1}{m} \sum_{i=1}^{m} \ell\left(y_i, h_{\mathcal{A}}(x_i)\right) - \frac{1}{u} \sum_{j=m+1}^{m+u} \ell\left(y_i, h_{\mathcal{A}}(x_i)\right) \right| \\
&= \left| R_{h_{\mathcal{A}}}(X_m) - R_{h_{\mathcal{A}}}(X_u) \right|.
\end{aligned}
$$

Consider Setting 2. The probability that $C_{\mathcal{A}}$ deviates from zero by an amount greater than $\varepsilon$ is

$$
P = \int_{\mathcal{X}, \mathcal{Y}} \mathbb{I}(C_{\mathcal{A}} - \varepsilon) d\mu(x_1, y_1) \cdots d\mu(x_{m+u}, y_{m+u}),
$$

where $\mathbb{I}$ is an indicator step function, $\mathbb{I}(x) = 1$ iff $x > 0$ and $\mathbb{I}(x) = 0$ otherwise. Let $\mathcal{T}_p$, $p = 1, \ldots, (m+u)!$ be the permutation operator for the sample $(x_1, y_1); \ldots; (x_{m+u}, y_{m+u})$. It is not hard to see that

$$
P = \int_{\mathcal{X}, \mathcal{Y}} \left\{ \frac{1}{(m+u)!} \sum_{p=1}^{(m+u)!} \mathbb{I}\left(C_{\mathcal{A}}(\mathcal{T}_p(x_1, y_1; \ldots; x_{m+u}, y_{m+u})) - \varepsilon\right) \right\} d\mu(x_1, y_1) \cdots d\mu(x_{m+u}, y_{m+u}).
$$

The expression in curly braces is the quantity estimated in Setting 1 and by our assumption, for any choice of the full sample, it does not exceed $\delta$. Therefore,

$$
P \leq \int_{\mathcal{X}, \mathcal{Y}} \delta d\mu(x_1, y_1) \cdots d\mu(x_{m+u}, y_{m+u}) = \delta.
$$

■

## References

Banerjee, A., & Langford, J. (2004). An objective evaluation criterion for clustering. Tenth ACM SIGKDD International Conference on Knowledge Discovery and Data Mining (KDD). Available at: `http://hunch.net/~jl/projects/prediction_bounds/clustering/clustering.ps`.

Baram, Y., El-Yaniv, R., & Luz, K. (2004). Online choice of active learning algorithms. *Journal of Machine Learning Research*, *5*, 255–291.

Bennett, K., & Demiriz, A. (1998). Semi-supervised support vector machines. *Advances in Neural Information Processing Systems*, *12*, 368–374.

Blum, A., & Langford, J. (2003). PAC-MDL Bounds. In *Proceedings of the Sixteenth Annual Conference on Computational Learning Theory*, pp. 344–357.

Bottou, L., Cortes, C., & Vapnik, V. (1994). On the effective VC dimension.. Tech. rep. bottou-effvc.ps.Z, Neuroprose (ftp://archive.cis.ohio-state.edu/pub/neuroprose). Also on http://leon.bottou.com/publications.

Boucheron, S., Lugosi, G., & Massart, P. (2003). Concentration inequalities using the entropy method. *The Annals of Probability*, *31*, 1583–1614.

Cover, T., & Thomas, J. (1991). *Elements of Information Theory*. John Wiley & Sons, New York.

Dembo, A., & Zeitouni, O. (1998). *Large Deviation Techniques and Applications* (Second edition). Springer, New York.

Demiriz, A., & Bennett, K. (2000). Optimization approaches to semi-supervised learning. In Ferris, M., Mangasarian, O., & Pang, J. (Eds.), *Complementarity: Applications, Algorithms and Extensions*, Vol. 50 of *Applied Optimization*, chap. 1, pp. 1–19. Kluwer.

El-Yaniv, R., & Gerzon, L. (2004). Effective transductive learning via PAC-Bayesian model selection. Tech. rep. CS-2004-05, Technion - Israel Institute of Technology. Available at `http://www.cs.technion.ac.il/Research/TechnicalReports/index.html`.

Herbrich, R. (2002). *Learning Kernel Classifiers: Theory and Algorithms*. MIT Press, Boston.

Hoeffding, W. (1963). Probability inequalities for sums of bounded random variables. *J. Amer. Statis. Assoc.*, *58*, 13–30.

Hush, D., & Scovel, C. (2003). Concentration of the hypergeometric distribution. Tech. rep. LA-UR-03-1353, Los Alamos National Laboratory.

Joachims, T. (1999). Transductive inference for text classification unsing support vector machines. In *European Conference on Machine Learning*.

Lanckriet, G., Cristianini, N., Ghaoui, L. E., Bartlett, P., & Jordan, M. (2002). Learning the kernel matrix with semi-definite programming. Tech. rep., University of Berkeley, Computer Science Division.

Langford, J., & Shawe-Taylor, J. (2002). PAC-Bayes and margins. In *Advances in Neural Information Processing Systems (NIPS 2001)*, pp. 439–446.

Lugosi, G. (2003). Concentration-of-measure inequalities. Tech. rep., Department of Economics, Pompeu Fabra University. Machine Learning Summer School 2003, Australian National University. See http://www.econ.upf.es/ lugosi/anu.ps.

McAllester, D. (1999). Some PAC-Bayesian theorems. *Machine Learning*, *37(3)*, 355–363.

McAllester, D. (2003a). PAC-Bayesian stochastic model selection. *Machine Learning*, *51(1)*, 5–21.

McAllester, D. (2003b). Simplified PAC-Bayesian margin bounds. In *COLT*, pp. 203–215.

McAllester, D. A. (1999). PAC-Bayesian model averaging. In *Proceedings of the twelfth Annual conference on Computational learning theory*, New York. ACM Press.

McDiarmid, C. (1989). On the method of bounded differences. In *Surveys in Combinatorics*, pp. 148–188. Cambridge University Press.

Seeger, M. (2002). Learning with labeled and unlabeled data. Tech. rep., Available at http://www.dai.ed.ac.uk/homes/seeger/papers/.

Seeger, M. (2003). *Bayesian Gaussian process models: PAC-Bayesian generalization error bounds and sparse approximations*. Ph.D. thesis, University of Edinburgh, Edinburgh.

Serfling, R. (1974). Probability inequalities for the sum in sampling without replacacement. *The Annals of Statistics*, *2*(1), 39–48.

Vapnik, V. N. (1982). *Estimation of Dependences Based on Empirical Data*. Springer Verlag, New York.

Vapnik, V. N. (1995). *The Nature of Statistical Learning Theory*. Springer Verlag, New York.

Vapnik, V. N. (1998). *Statistical Learning Theory*. Wiley Interscience, New York.

Wu, D., Bennett, K., Cristianini, N., & Shawe-Taylor, J. (1999). Large margin trees for induction and transduction. In *International Conference on Machine Learning*.